\documentclass{article}
\PassOptionsToPackage{numbers, compress}{natbib}
\usepackage[preprint]{neurips_2020}
\usepackage{amssymb} % lets one use \mathbb
\usepackage{amsmath}
\usepackage{amsthm}
\usepackage{mathtools}
\usepackage[breaklinks=true]{hyperref}
\usepackage{cleveref}

\usepackage{xcolor}
\usepackage[utf8]{inputenc}
\usepackage[T1]{fontenc}

\usepackage{textcomp} % otherwise bibtex breaks

\usepackage{graphicx}

\usepackage[caption=false]{subfig}

\newcommand{\norm}[1]{\left\lVert#1\right\rVert}
\newcommand{\ltwonorm}{\(\ell_2\)-norm }
\newcommand{\lonenorm}{\(\ell_1\)-norm }
\DeclareMathOperator{\tensordotvec}{\bar{\times}}

\title{Deep convolutional tensor network}
\author{%
  Philip Blagoveschensky \\
  Skolkovo Institute of Science and Technology \\
  Moscow, Russia 121205 \\
  \texttt{philip-b@crabman.me} \\
  \And
  Anh-Huy Phan \\
  Skolkovo Institute of Science and Technology \\
  Moscow, Russia 121205 \\
  \texttt{a.phan@skoltech.ru} \\
}

\begin{document}

\maketitle

\begin{abstract}
  Neural networks have achieved state of the art results in many areas, supposedly due to
  parameter sharing, locality, and depth. Tensor networks (TNs) are linear algebraic representations
  of quantum many-body states based on their entanglement structure. TNs have found use in
  machine learning. We devise a novel TN based model called Deep
  convolutional tensor network (DCTN) for image classification, which has parameter sharing,
  locality, and depth. It is based on the Entangled plaquette states (EPS) TN. We
  show how EPS can be implemented as a backpropagatable layer.
  We test DCTN on MNIST, FashionMNIST, and CIFAR10 datasets. A
  shallow DCTN performs well on MNIST and FashionMNIST and has a small parameter
  count. Unfortunately, depth increases overfitting and thus decreases test accuracy. Also,
  DCTN of any depth performs badly on CIFAR10 due to overfitting. It is to be determined
  why. We discuss how the hyperparameters of DCTN affect its training and overfitting.
\end{abstract}

\section{Introduction}
\label{sec:introduction}
\subsection{Properties of successful neural networks}
\label{sec:properties}

Nowadays, neural networks (\emph{NNs}) achieve outstanding results in many machine learning tasks~\citep{paperswithcode_sota}, including computer vision, language modeling, game playing (e.g. Checkers, Go), automated theorem proving~\citep{gpt_f}. There are three properties many (but not all) NNs enjoy, which are thought to be responsible for their success. For example, \citep{cohen2016expressive} discusses the importance of these properties for deep CNNs.

\begin{itemize}
\item \emph{Parameter sharing}, aka applying the same transformation multiple times in parallel or sequentially. A layer of a convolutional neural network (\emph{CNN}) applies the same function, defined by a convolution kernel, to all sliding windows of an input. A recurrent neural network (RNN) applies the same function to the input token and the hidden state at each time step. A self-attention layer in a transformer applies the same query-producing, the same key-producing, and the same value-producing function to each token.~\citep{illustrated_transformer}
\item \emph{Locality}. Interactions between nearby parts of an input are modeled more accurately, while interactions between far away parts are modeled less accurately or not modeled at all. This property makes sense only for some types of input. For images, this is similar to receptive fields in a human's visual cortex. For natural language, nearby tokens are usually more related than tokens far away from each other. CNNs and RNNs enjoy this property.
\item \emph{Depth}. Most successful NNs, including CNNs and transformers, are
deep, which allows them to learn complicated transformations.
\end{itemize}

\subsection{The same properties in tensor networks}

Tensor networks (\emph{TNs}) are linear algebraic representations of quantum many-body states based on
their entanglement structure. They've found applications in signal processing.
People are exploring their applications to machine
learning, e.g. tensor regression -- a class of machine learning models
based on contracting (connecting the edges) an input tensor with a parametrized TN. Since NNs with the
three properties mentioned in \Cref{sec:properties} are so successful, it would make sense to try to devise a
tensor regression model with the same properties. That is what we do in our paper. As far as we
know, some existing tensor networks have one or two out of the three properties, but none have
all three.
\begin{itemize}
\item MERA (see Ch. 7 of \citep{bridgeman2017handwaving_and_interpretive_dance}) is a
  tree-like tensor network used in quantum many-body physics. It's deep and has locality.
\item Deep Boltzmann machine can be viewed as a tensor network. (See Sec. 4.2 of
  \citep{cichocki_part_2} or \citep{glasser2018probabilistic} for discussion of how
  restricted Boltzmann machine is actually a tensor network. It's not difficult to see a DBM
  is a tensor network as well). For supervised learning, it can be viewed as tensor
  regression with depth, but without locality or weight sharing.
\item \citep{glasser2018probabilistic} introduced Entangled plaquette states (\emph{EPS}) with
  weight sharing for tensor regression. They combined one EPS with a linear classifier or a
  matrix tensor train. Such a model has locality and parameter sharing but isn't deep.
\item \citep{cohen2016expressive} introduced a tensor regression model called Deep
  convolutional arithmetic circuit. However, they used it only theoretically to analyze the
  expressivity of deep CNNs and compare it with the expressivity of tensor regression
  with tensor
  in CP format (canonical polyadic / CANDECOMP PARAFAC).  Their main result is a theorem about
  the typical canonical rank of a tensor network used in Deep convolutional arithmetic
  circuit. The tensor network is very similar to the model we propose, with a few small
  modifications. We conjecture that the proof of their result about the typical canonical rank
  being exponentially large can be modified to apply to our tensor network as well.
\item \citep{miller_2020_umps_psm} did language modeling by contracting an input sequence with
  a matrix tensor train with all cores equal to each other. It has locality and parameter
  sharing.
\item \citep{liu2019ml_by_unitary_tn_of_hierarchical_tree_structure} used a tree-like
  tensor regression model with all cores being unitary. Their model has locality and depth,
  but no weight sharing.
\item \citep{stoudenmire1605supervised} and \citep{novikov2016exponential} performed
  tensor regression on MNIST images and tabular datasets, respectively. They encoded input data
  as rank-one tensors like we do in \Cref{sec:input_preprocessing} and contracted it with a matrix
  tensor train to get predictions. Such a model has locality if you order the matrix tensor
  train cores in the right way.
\end{itemize}

\subsection{Contributions}

The main contributions of our article are:
\begin{itemize}
\item We devise a novel tensor regression model called Deep convolutional tensor network
  (\emph{DCTN}). It has all three properties listed in \Cref{sec:properties}. It is based on
  the (functional) composition of TNs called Entangled plaquette state (\emph{EPS}).
  DCTN is similar to a deep CNN. We apply it to image classification, because that's the
  most straightforward application of deep CNNs. (\Cref{sec:description_of_the_whole_model})
\item We show how EPS can be implemented as a
  backpropagatable function/layer which can be used in neural networks or other
  backpropagation based models (\Cref{sec:entangled_plaquette_states}). 
\item Using common techniques for training deep neural networks, we train and evaluate DCTN on
  MNIST, FashionMNIST, and CIFAR10 datasets. A shallow model based on one EPS works well on
  MNIST and FashionMNIST and has a small parameter count. Unfortunately, increasing depth of
  DCTN by adding more EPSes hurts its accuracy by increasing overfitting. Also, our model works
  very badly on CIFAR10 regardless of depth. We discuss hypotheses why this is the
  case. (\Cref{sec:experiments}).
\item We show how various hyperparameters affect the model's optimization and generalization
  (\Cref{sec:how_hyperparameters_affect_optimization_and_generalization}).
\end{itemize}

\section{Notation}
\begin{figure}[h]
  \centering
  \includegraphics[width=0.7\textwidth]{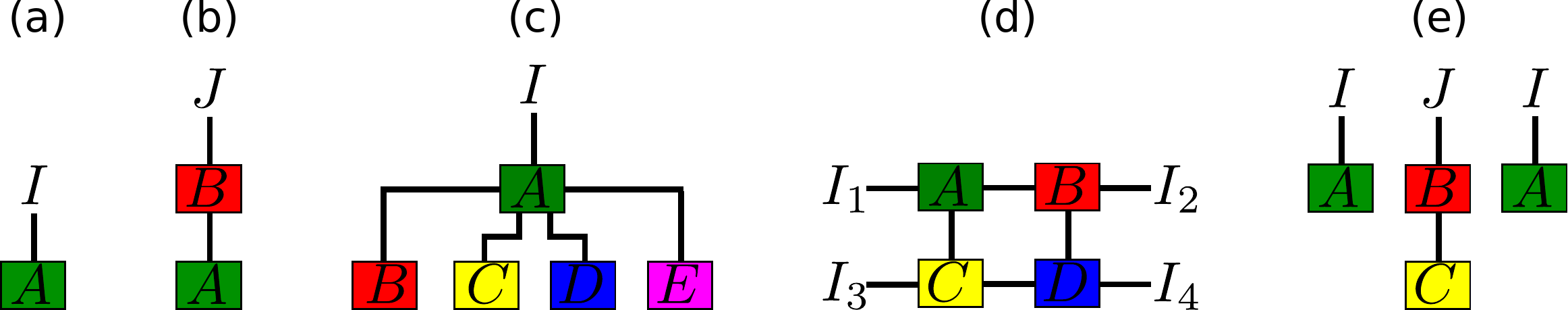}
  \caption[TN diagrams]{(a) Vector \(A \in \mathbb{R}^I\). (b) Matrix-vector
    product \(BA = B \tensordotvec_2 A \in \mathbb{R}^J\). (c) Contraction of an order-5 tensor \(A\)
    with 4 vectors \(B, C, D, E\) results in vector \(A \tensordotvec_5 E \tensordotvec_4 D \tensordotvec_3 C
    \tensordotvec_2 B \in \mathbb{R}^I\). (d) Contraction of four order-3 tensors, which results in an
  order-4 tensor. (e) Traditionally, if a TN diagram contains disconnected subgraphs, the whole
  diagram represents their outer product \(A \otimes BC \otimes A \in \mathbb{R}^{I \times J
    \times I}\). However, in our work, we usually interpret it as simply a collection of
  tensors \((A, BC, A)\).}
  \label{fig:tn}
\end{figure}

An order-\(N\) tensor is a real valued multidimensional array
\(A \in \mathbb{R}^{I_1 \times \dots \times I_N}\). A scalar is an order-0
tensor, a vector is an order-1 tensor, a matrix is an order-2 tensor. We refer to a slice of a
tensor using parentheses, e.g.
\(A(i_1, \dots, i_k) \in \mathbb{R}^{I_{k+1} \times \dots \times I_N}\),
\(A(i_1, \dots, i_N) \in \mathbb{R}\). In the second case we got a scalar, because we fixed all
indices. We can vectorize a tensor to
get \(\operatorname{vec}(A) \in \mathbb{R}^{I_1 \cdots I_N}\). If a tensor has order at
least 2, we can separate its dimensions into two disjoint sets \(L \sqcup R = \{I_1, \dots,
I_N\}\) and matricize the tensor, i.e. turn it into an
\(\prod_{I \in L} I\) by \(\prod_{I \in R} I\) matrix \(\operatorname{mat}(A)\). When we use
matricization, we won't explicitly specify how we separate the dimensions into disjoint
sets, because it should be clear from context.
If we also have a tensor
\(B \in \mathbb{R}^{J_1 \times \dots \times J_M}\), their outer product \(A \otimes B \in
\mathbb{R}^{I_1 \times \dots \times I_N \times J_1 \times \dots \times J_M}\) is defined as
\[(A \otimes B)(i_1, \dots, i_N, j_1, \dots, j_M) = A(i_1, \dots, i_N) B(j_1, \dots, j_M).\]
If, in addition to \(A\), for some \(n \in \{1, \dots, N\}\) we have a vector
\(x \in \mathbb{R}^{I_n}\), we can
contract them on the \(n\)-th dimension of \(A\) to produce an order-\((N-1)\) tensor
\(A \tensordotvec_n x \in \mathbb{R}^{I_1 \times \dots \times I_{n-1} \times I_{n+1} \times I_N}\)
defined as
\[(A \tensordotvec_n x)(i_1, \dots, i_{n-1}, i_{n+1}, \dots, i_N) = \sum_{i_n=1}^{I_n} A(i_1, \dots,
  i_{n-1}, i_n, i_{n+1}, \dots, i_N) x(i_n).\] Contraction can also be performed between a
tensor and a tensor. However, to denote this we use TN
diagrams instead of formulas -- see \Cref{fig:tn}.

There are multiple introductions to TN
diagrams available. We recommend Chapter 1 of
\citep{bridgeman2017handwaving_and_interpretive_dance},
\citep{ehrbar2000graph_notation_for_arrays} (this article doesn't call them TNs,
but they are), and Chapter 2 of \citep{cichocki2016part1}. Other introductions, which are less
accessible for machine learning practitioners, are
\citep{orus2014practical_introduction_to_tns} and
\citep{biamonte2017tensor_networks_in_a_nutshell}.

\begin{figure}[h]
\centering
\subfloat[]{\includegraphics[scale=0.36]{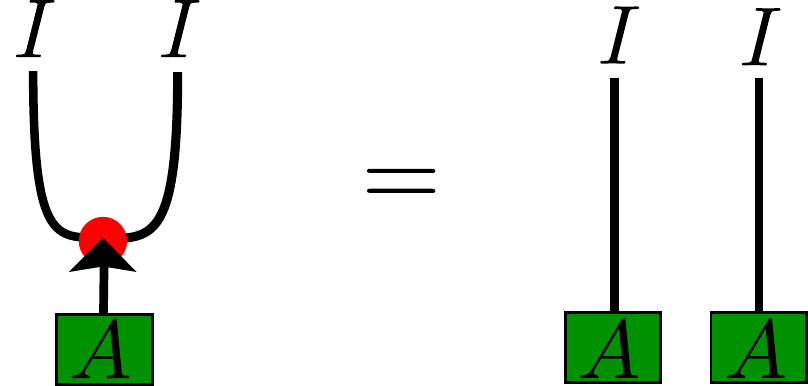}}
\subfloat[]{\includegraphics[scale=0.36]{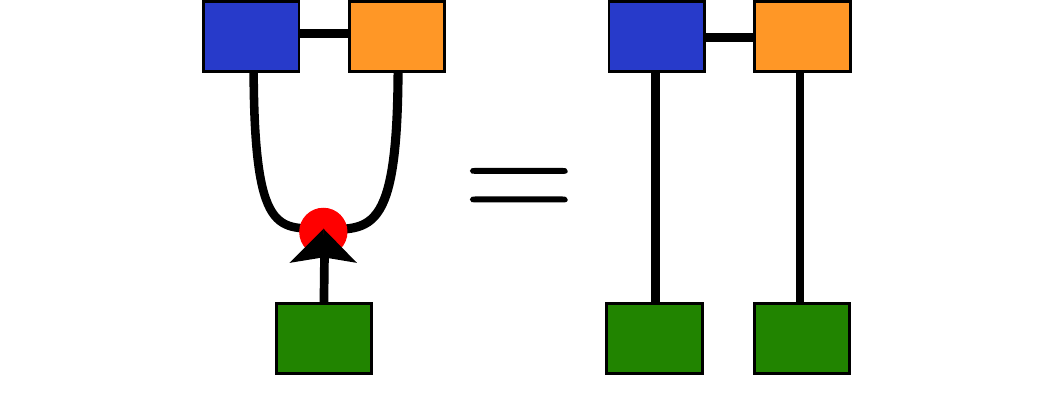}}
\subfloat[]{\includegraphics[scale=0.36]{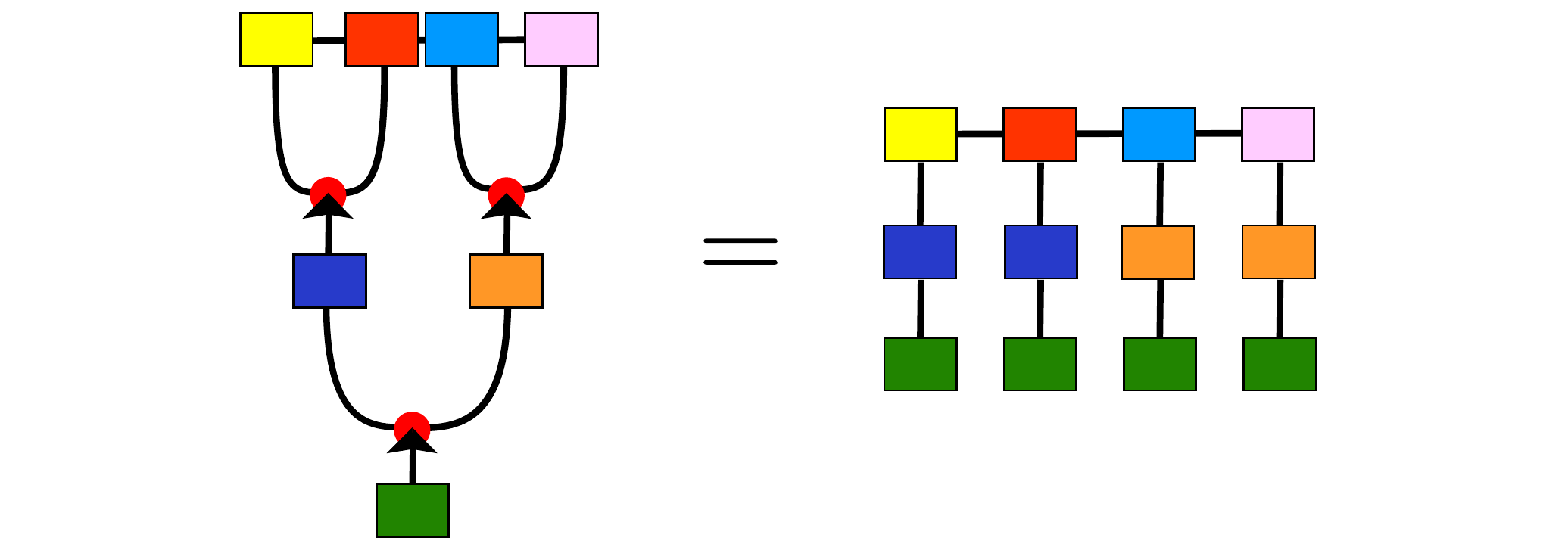}}\\
\subfloat[]{\includegraphics[scale=0.36]{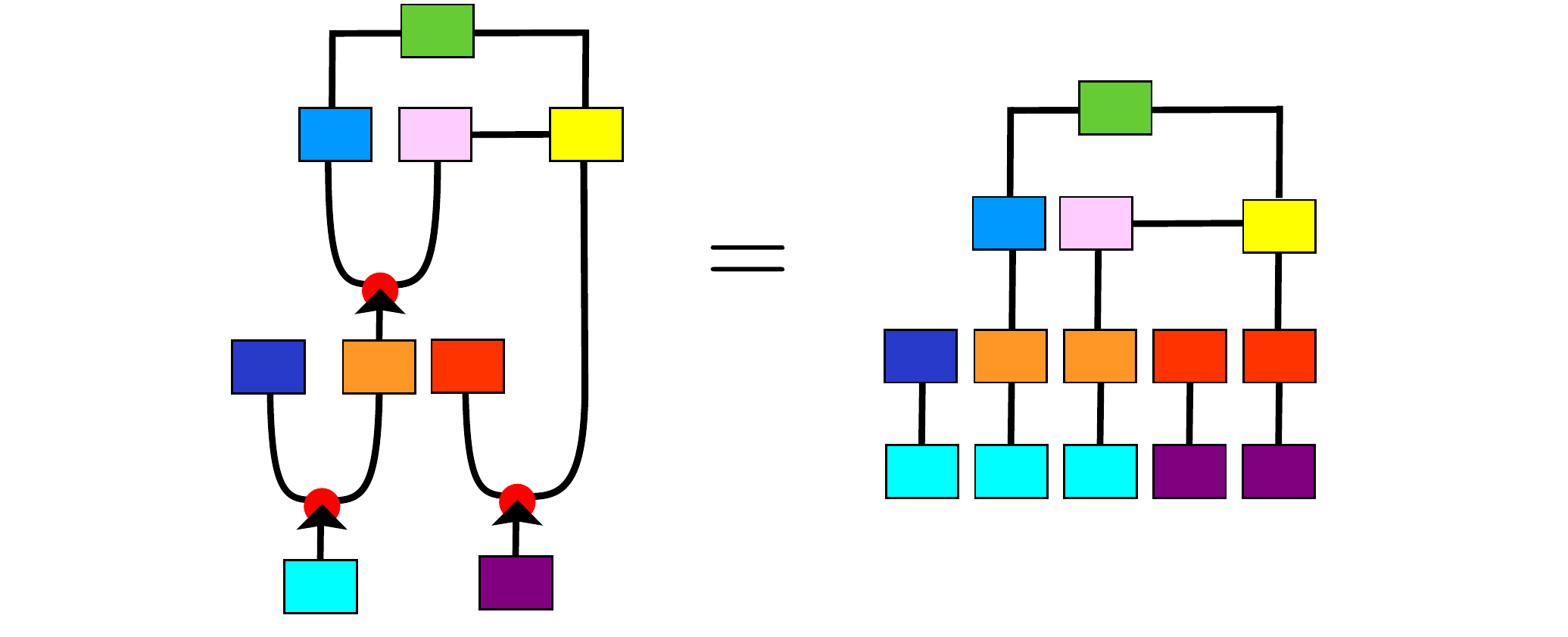}}
\subfloat[]{\includegraphics[scale=0.36]{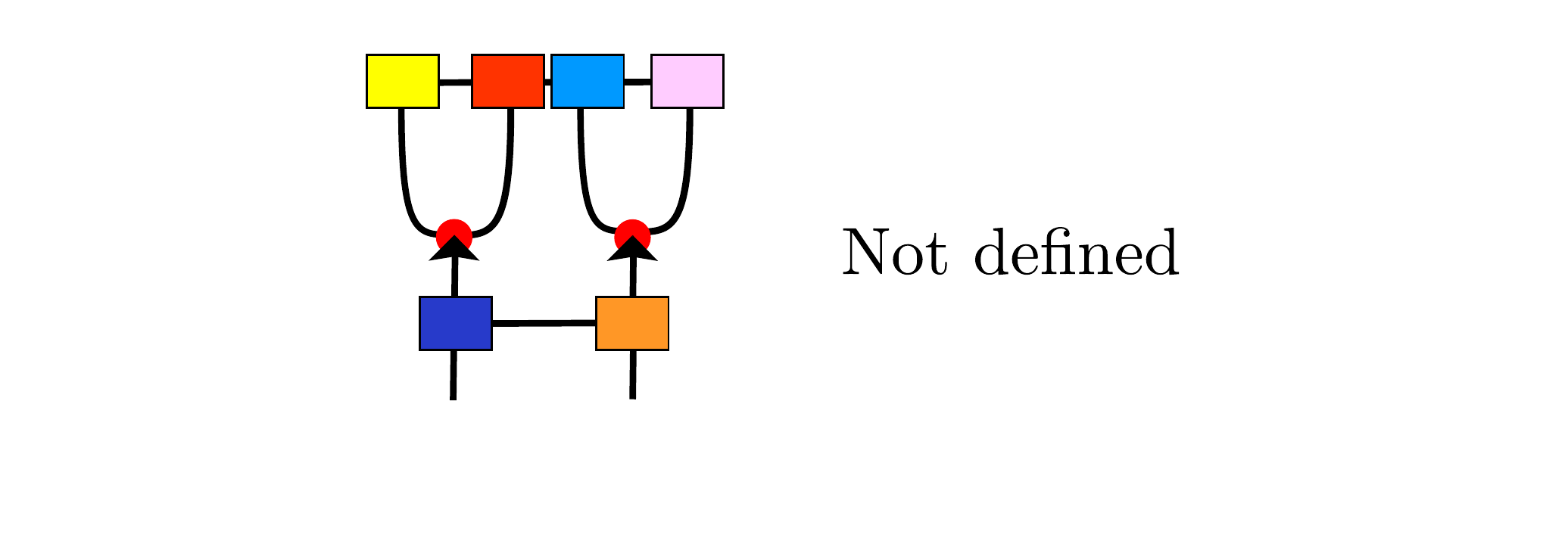}}%
\caption[Copy operation]{(a) Copy operation of a vector input $A$. The result can be
  interpreted either as the matrix $A \otimes A$ or as just a collection of two vectors
  \((A, A)\).
  (b), (c) and (d) : Mapping of generalized TNs to TNs with parameter sharing
  (tensors with the same colors are identical).
  (e) This representation is not defined because we define the copy
  operation only for vectors.
  Adapted from \citep{glasser2018probabilistic}. Copyright 2020 by
  Ivan Glasser, Nicola Pancotti, and J. Ignacio
  Cirac. \href{https://creativecommons.org/licenses/by/4.0/}{CC BY 4.0}}
\label{fig:copy_operation}
\end{figure}

We extend the notion of TNs
by introducing the \emph{copy operation}. We call tensor networks with the copy
operation \emph{generalized TNs}. The copy operation was invented by
\citep{glasser2018probabilistic}. It takes a vector as input and outputs
multiple copies of that vector. In generalized TN diagrams, we graphically depict
the copy operation by a red dot with one input edge marked with an arrow, and all output edges
not marked in any way. This operation is equivalent to having multiple copies of the input
contracted with the rest of the tensor network. In order for a generalized tensor network to be
well defined, it must have no directed cycles going through copy elements (if we consider the
usual edges between tensors to have both directions). \Cref{fig:copy_operation} explains the
copy operation in more detail.

\section{DCTN's description}
\subsection{Input preprocessing}
\label{sec:input_preprocessing}
\begin{figure}[h]
  \centering
  \includegraphics[scale=0.44]{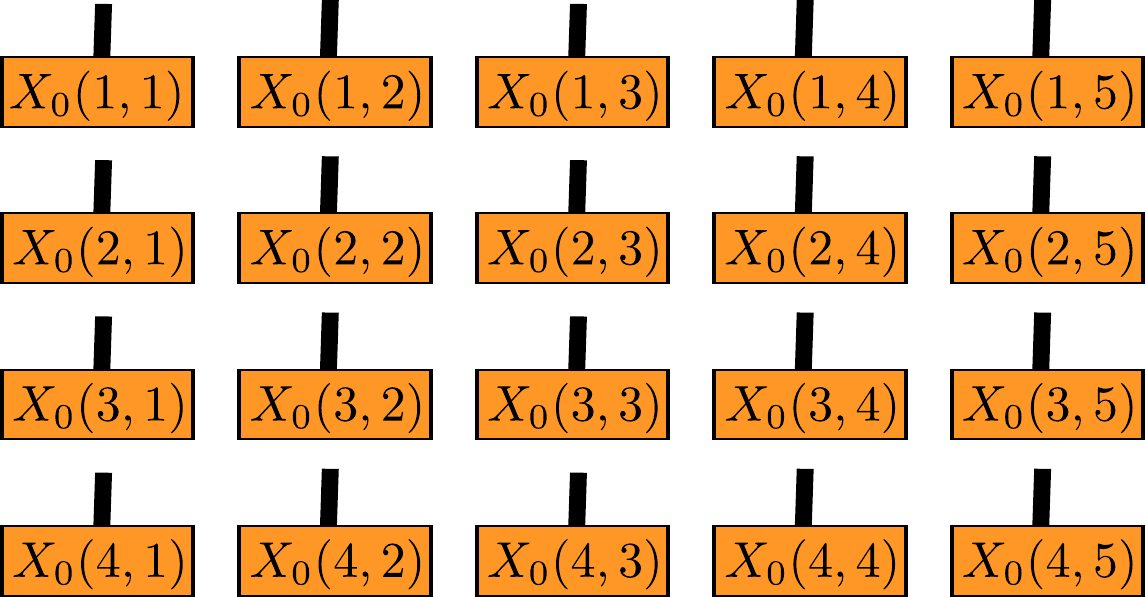}  
  \caption{The 0-th representation of a 4 by 5 image as a TN diagram.  We think of this TN not
    as an order-20 rank-1 tensor, but as just a collection of 20 vectors arranged on a grid.}
  \label{fig:image_as_rank_one_tensor}
\end{figure}

Suppose we want to classify images with height \(H\) pixels, width \(W\) pixels, and each pixel
is encoded with \(C\) color channels, each having a number in \([0, 1]\). Such an image is usually represented as a
tensor \(X \in [0, 1]^{H \times W \times C}\). We want to represent it in another way. We will
call this other way the 0th representation \(X_0\) of the image, meaning that it's the
representation before the first layer of DCTN. Throughout our work, when we
will be using variables to talk about the zeroth representation of an image, we will be giving
the variables names with the subscript 0, and for other representations we will be giving
variables names with other subscripts. We denote \(H_0=H\), \(W_0=W\), so for some small
positive integer \(Q_0\) (where \(Q\) stands for ``quantum dimension''),
we want to represent the image as \(HW\) vectors, each of size \(Q_0\):
\[\forall h \in \{1, \dots, H\} \, \forall w \in \{1, \dots, W\} \
  X_0(h, w) = \varphi(X(h, w)) \in \mathbb{R}^{Q_0},\]
where \(\varphi: [0, 1]^C \to \mathbb{R}^{Q_0}\) is some vector-valued function.
Such representation of an image constitutes a TN
with \(H_0 W_0\) vectors, none of them connected.
See \Cref{fig:image_as_rank_one_tensor} for illustration.

For grayscale images, i.e. \(C=1\), we set \(Q_0=2\). In this case we can omit
the dimension \(C\). We must choose \(\varphi: [0, 1] \to \mathbb{R}^{Q_0}\).
Possible choices include:
(1) \(\varphi(x) = \begin{bmatrix} \cos(\frac{\pi}{2} x) \\ \sin(\frac{\pi}{2} x) \end{bmatrix}\)
\citep{grant2018hierarchical,bhatia2019matrix,stoudenmire1605supervised,huggins2019towards}
encodes the value \(x\) in a qubit. The components are nonnegative. It has \ltwonorm equal to 1.
(2) \(\varphi(x) = \begin{bmatrix} x \\ 1-x \end{bmatrix}\) \citep{torchmps}. The components are
nonnegative. It has
\lonenorm equal to 1.
(3) \(\varphi(x) = \begin{bmatrix} \cos^2(\frac{\pi}{2} x) \\ \sin^2(\frac{\pi}{2} x)
\end{bmatrix}\) \citep{glasser2018probabilistic}. The components are nonnegative. The authors
say that the \lonenorm always being equal to 1 provides numerical stability.

% If I need more space, I can remove the next sentence
In light of the duality of tensor networks and discrete undirected probabilistic graphical
models \citep{robeva2019duality,glasser2018probabilistic}, the second and third choices can be
viewed as encoding a number as a binary probability distribution. In our work, we use
\begin{equation}
  \label{eq:phi_definition}
  \varphi(x) = \nu \begin{bmatrix}\cos^2(\frac{\pi}{2} x) \\ \sin^2(\frac{\pi}{2}
    x)\end{bmatrix},
\end{equation}
where \( \nu \) is some positive real number. The choice of \(\nu\) is described in \Cref{sec:initialization_and_scaling_of_input}.

When working with a colored dataset, we convert the images to YCbCr,
normalize, and add a fourth channel of constant ones. In other words, we use
\[ \phi\left(\begin{bmatrix}y & b & r\end{bmatrix}^T\right)
  = \begin{bmatrix}\frac{y-\mu_y}{\sigma_y} & \frac{b-\mu_b}{\sigma_b} &
    {r-\mu_r}{\sigma_r} & 1\end{bmatrix}^T,\]
where \(\mu_y, \mu_b, \mu_r, \sigma_y, \sigma_b, \sigma_r\) are means and standard deviations
(over the training dataset) of the three channels Y, Cb, Cr, correspondingly.

\subsection{Entangled plaquette states}
\label{sec:entangled_plaquette_states}
Entangled plaquette states (\emph{EPS}) is defined in \citep{glasser2018probabilistic} as a
generalized TN, in which vectors arranged on a two-dimensional grid are contracted with tensors
of parameters. Suppose \(K\) is a small positive integer called the \emph{kernel size}
(having the same meaning as kernel size in \emph{Conv2d} function). Suppose \(Q_{\text{in}},
Q_{\text{out}}\) are positive integers called the \emph{quantum dimension size of input} and
the \emph{quantum dimension size of output}, respectively.
Then an EPS is
parametrized with an order-\((K^2+1)\) tensor
\(E \in \mathbb{R}^{Q_{\text{out}} \times Q_{\text{in}} \times \dots \times Q_{\text{in}}}\)
% I can remove the next line if needed for SIZE
with one dimension of size \(Q_{\text{out}}\) and \(K^2\) dimensions of size \(Q_{\text{in}}\).

\begin{figure}[h]
  \centering
  \subfloat[]{\includegraphics[scale=0.5]{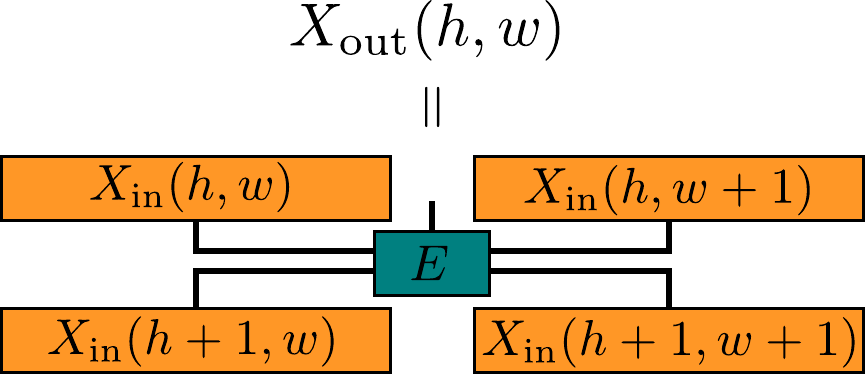}}
  \qquad
  \subfloat[]{\includegraphics[scale=0.5]{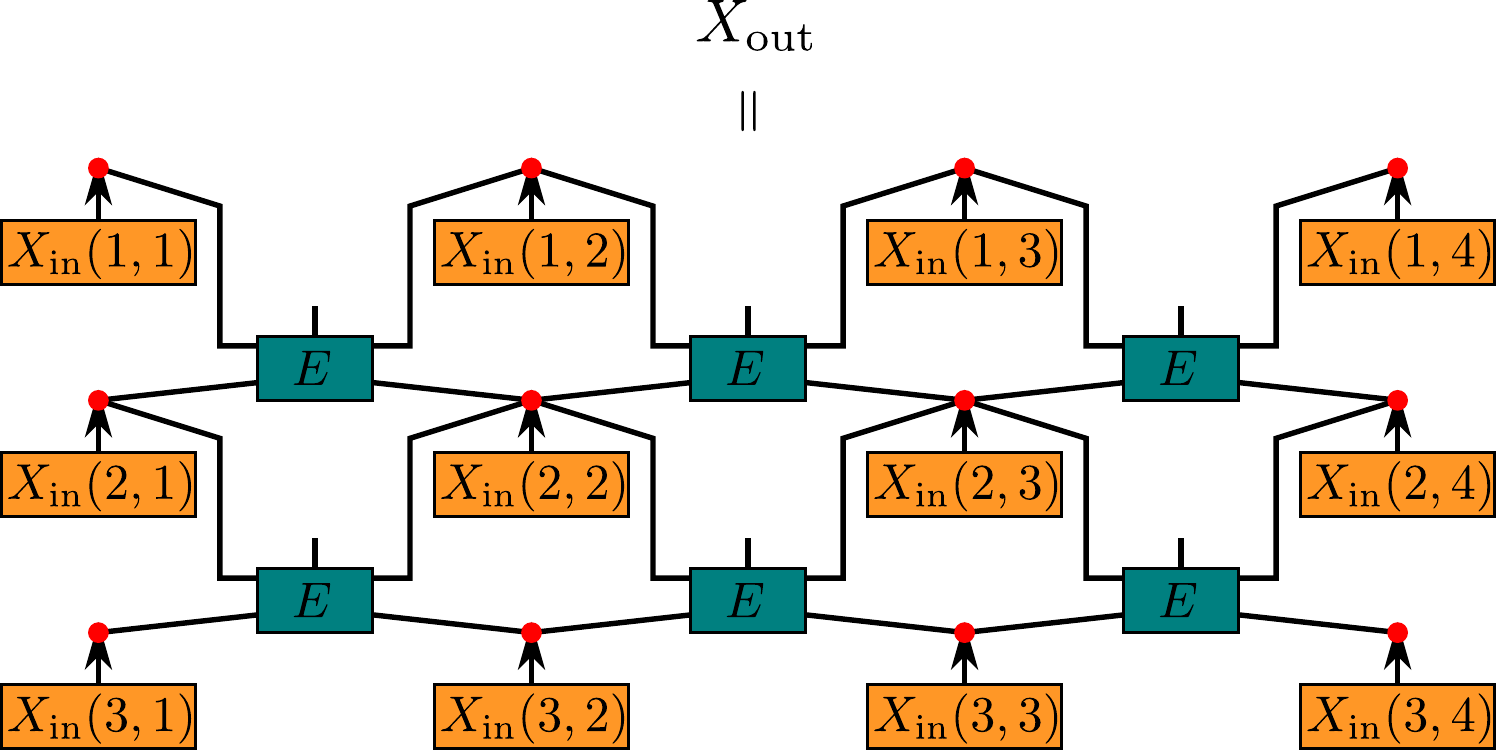}}%
  \caption{(a) One pixel of the output of an EPS with kernel size \(K=2\) visualized as a TN
    diagram. The TN is equivalent to
    \cref{eq:eps_output_one_pixel_contractions,eq:eps_output_one_pixel_matmul}.
    (b) An EPS applied to an input is a generalized TN. Here \(H_{\text{in}}=3,
    W_{\text{in}}=4, K=2\). Teal cores are identical.
    The output \(X_{\text{out}}\) is a collection of 6 vectors.
    According to the terminology of \citep{miller_2020_umps_psm}, an EPS is a uniform TN.}
  \label{fig:eps}
\end{figure}

Suppose \(H_{\text{in}} \geq K, W_{\text{in}} \geq K\) are integers denoting the height and
width of an input \(X_{\text{in}}\) consisting of \(H_{\text{in}} W_{\text{in}}\) vectors
\(X_\text{in}(h, w) \in \mathbb{R}^{Q_{\text{in}}}\) arranged on a \(H_{\text{in}}\) by
\(W_{\text{in}}\) grid.
Then applying the EPS parametrized by \(E\) to the input \(X_\text{in}\) produces an output
\(X_\text{out} = \operatorname{eps}(E, X_\text{in})\) consisting of vectors
\(\operatorname{eps}(E, X_\text{in})(h, w) \in \mathbb{R}^{Q_{\text{out}}}\)
arranged on a \(H_\text{out} = H_\text{in} - K + 1\) by 
\(W_\text{out} = W_\text{in} - K + 1\) grid defined
as
\begin{equation}
  X_\text{out}(h, w) = E
   \bar{\times}_{K^2+1} X_\text{in}(h+0, w+0) \bar{\times}_{K^2} X_\text{in}(h+0, w+1) \dots
  \bar{\times}_{2} X_\text{in}(h+K-1, w+K-1).
  \label{eq:eps_output_one_pixel_contractions}
\end{equation}
\Cref{fig:eps} visualizes this formula and shows that an EPS applied to an input is a
generalized TN.
Using matricization
\(\operatorname{mat}(E) \in \mathbb{R}^{Q_\text{out} \times Q_\text{in}^{K^2}}\),
this formula can be rewritten as
\begin{equation}
  X_\text{out}(h, w) = \operatorname{mat}(E) \cdot
    \operatorname{vec}\left( \bigotimes_{h' = 0}^{K-1} \bigotimes_{w' = 0}^{K-1}
      X_\text{in}(h + h', w + w') \right).
    \label{eq:eps_output_one_pixel_matmul}
\end{equation}
Notice that application of an EPS to an input applies the same function to each \(K \times K\)
sliding window of the input. This provides two of the three properties described in
\Cref{sec:properties}: \emph{locality} and \emph{parameter sharing}.

EPS can be implemented as a backpropagatable function/layer and be used in neural networks or
other gradient descent based models. The formulas for forward pass are given in
\cref{eq:eps_output_one_pixel_matmul,eq:eps_output_one_pixel_contractions}.
Next, we provide the backward pass formulas for derivatives. For
\(\frac{\partial \operatorname{eps}(E, X_\text{in})(h, w)}{\partial X_\text{in}(h', w')}
\in \mathbb{R}^{Q_\text{out} \times Q_\text{in}}\),
if \(h' \in \{h, \dots, h+K-1\}\) and \(w' \in \{w,\dots,w+K-1\}\), we have
\[\frac{\partial \operatorname{eps}(E, X_\text{in})(h, w)}{\partial X_\text{in}(h', w')}=
E \underbrace{
  \tensordotvec_{K^2+1} X_{\text{in}}(h+K-1, w+K-1)
  \cdots                                                                                         
  \tensordotvec_3 X_{\text{in}}(h, w+1)
  \tensordotvec_2 X_{\text{in}}(h, w)
}_{\text{for each pair of indices } (h+\delta h, w+\delta w) \text{ except } (h+\delta h=h',w+\delta w=w')}
,\]
otherwise we have 
\(\frac{\partial \operatorname{eps}(E, X_\text{in})(h, w)}{\partial X_\text{in}(h', w')} = 0.\)
For
\( \frac{\partial \operatorname{eps}(E, X_\text{in})(h, w)}{\partial E}
  \in \mathbb{R}^{Q_\text{out} \times Q_\text{out} \times Q_\text{in} \times \dots \times
    Q_\text{in}}, \)
denoting \(I \in \mathbb{R}^{Q_\text{out} \times Q_\text{out}}\) to be the identity matrix, we have
\[
  \frac{\partial \operatorname{eps}(E, X_\text{in})(h, w)}{\partial E}
  = I \otimes \bigotimes_{\delta h = 0}^{K-1} \bigotimes_{\delta w = 0}^{K-1}
  X_\text{in}(h+\delta h, w + \delta w).
\]
% Here I also give specific formulas for people who don't care about tensor networks.

\subsection{Description of the whole model}
\label{sec:description_of_the_whole_model}
DCTN is a (functional) composition of the preprocessing function \( \varphi \)
described in \Cref{sec:input_preprocessing}, \(N\) EPSes described in
\Cref{sec:entangled_plaquette_states} parametrized by tensors \(E_1, \dots, E_N\), a linear
layer parametrized by a matrix \(A\) and a vector \(b\), and the softmax function. The whole
model is defined by

\begin{gather}
 X_0(h, w) = \varphi(X(h, w)) \label{eq:wholemodel1}\\
 X_1 = \operatorname{eps}(E_1, X_0) \label{eq:wholemodel2}\\
 X_2 = \operatorname{eps}(E_2, X_1) \label{eq:wholemodel3}\\
 \qquad  \vdots \notag\\
 X_N = \operatorname{eps}(E_N, X_{N-1}) \label{eq:wholemodel4}\\
 \ln \hat p(y=\ell \mid X) = (A \cdot \operatorname{vec}(X_N) + b)_\ell \label{eq:wholemodel5}\\
 p(y=\ell \mid X)
  = \operatorname{softmax}\left(
      \begin{bmatrix}
        \ln \hat p(y=1 \mid X) \\
        \ln \hat p(y=2 \mid X) \\
        \vdots \\
        \ln \hat p(y=L \mid X)
      \end{bmatrix}
    \right)_\ell
    = \frac{\hat p(y=\ell \mid X)}{\sum_{\ell'=1}^L \hat p(y=\ell' \mid X)}, \label{eq:wholemodel6}
\end{gather}
where \(L\) is the number of labels.
The original input image \(X\) is represented as a tensor of shape \(H_0 \times W_0 \times C\).
For each \(n\), the \(n\)-th intermediate representation \(X_n\) consists of \(H_n\) by \(W_n\)
vectors of size \(Q_n\), and it holds that \(H_n = H_{n-1} - K_n + 1\), \(W_n = W_{n-1} - K_n + 1\),
where \(K_n\) is the kernel size of the \(n\)-th EPS.
In principle, the affine function parametrized by \(A\) and \(b\) can be replaced with another
differentiable possibly parametrized function, for example another tensor network.

\begin{figure}[h]
  \centering
  \subfloat[]{\includegraphics[scale=0.6]{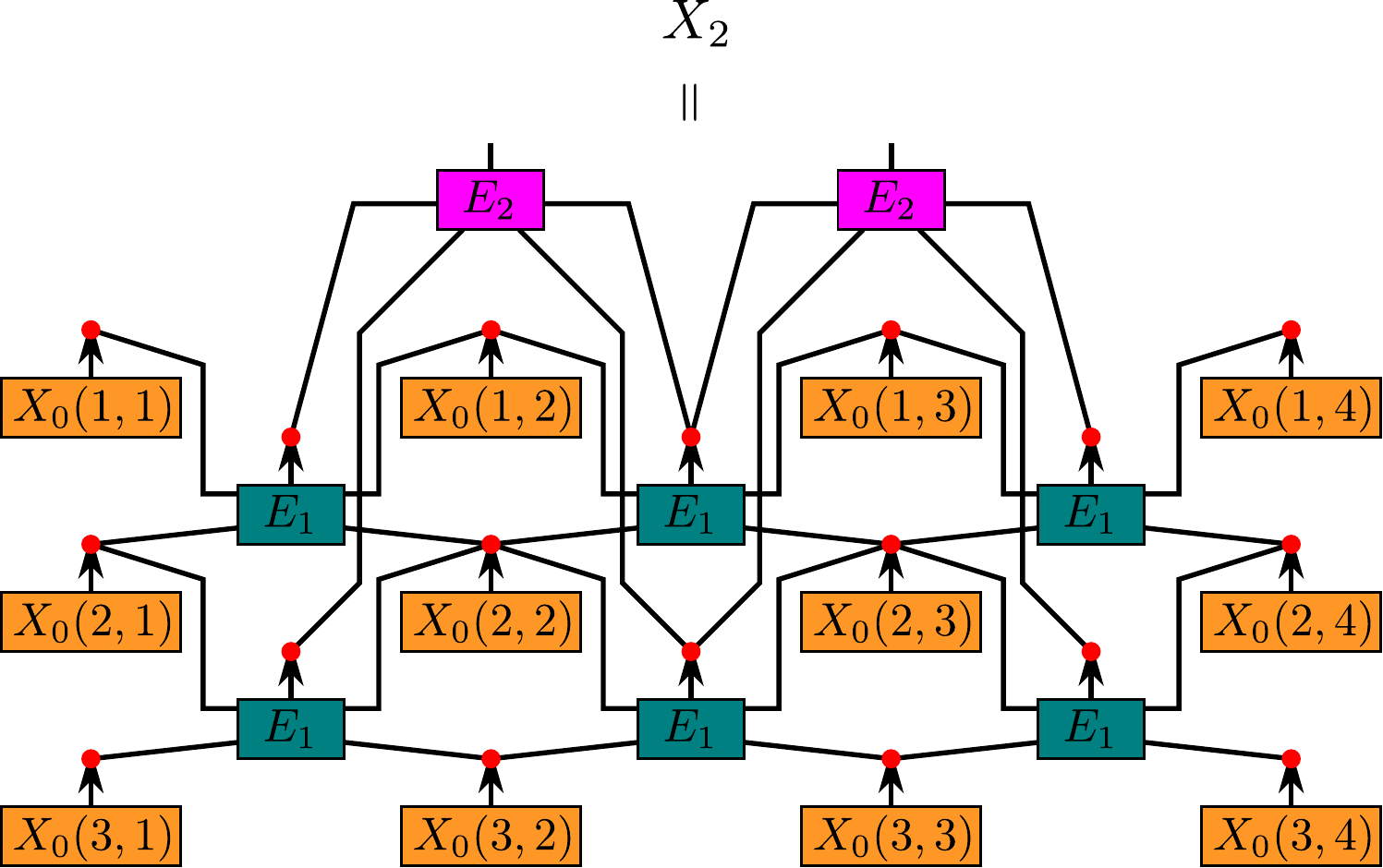}}
  \qquad
  \subfloat[]{\includegraphics[scale=0.6]{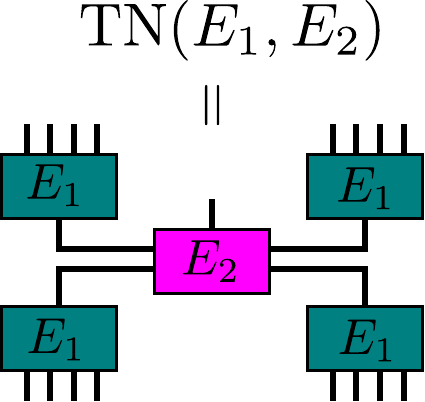}}
  \caption{A composition of EPSes \(E_1, E_2\) with \(K_1=K_2=2\). (a) A composition of EPSes
    applied to an input is a generalized TN. Here \(H_0=3, W_0=4\). (b) A composition of EPSes
    is a TN, denoted \(\operatorname{TN}(E_1, \dots, E_N)\). Here, \(\operatorname{TN}(E_1,
    E_2) \in \mathbb{R}^{Q_2 \times Q_0 \times Q_0 \times \dots \times Q_0}\) is an order-17
    tensor. If you contract this TN with with a sliding window of an input (with some pixels
    copied), you'll get one pixel of output.}
  \label{fig:two_epses}
\end{figure}

A composition of EPSes is a TN. A composition of EPSes applied to an input is a generalized
TN. See visualization in \Cref{fig:two_epses}.

\subsection{Optimization}
We initialize parameters \(E_1, \dots, E_N\) and \(A, b\) of DCTN randomly.
(\Cref{sec:initialization_and_scaling_of_input} contains details.)
Let \(\lambda \geq 0\) be the regularization coefficient. To train DCTN, at
each iteration we sample \(M\) images \(X^{(1)}, \dots, X^{(M)}\) and their labels
\(y^{(1)}, \dots, y^{(M)}\) from the training dataset and use Adam optimizer
\citep{kingma2014adam} with either the objective
\begin{equation}
  \label{eq:opt_problem_whole_tn_l2_reg}
  \underset{E_1, \dots, E_N, A, b}{\text{minimize}} \ \lambda \left(\norm{\operatorname{TN}(E_1,
      \dots, E_N)}_\text{fro}^2 + \norm{A}_\text{fro}^2 \right)
  + \frac{1}{M} \sum_{m=1}^M -\ln p\left( y^{(m)} \mid X^{(m)}\right),
\end{equation}
where \(\operatorname{TN}(E_1, \dots, E_N)\) is defined in \Cref{fig:two_epses},
or the objective
\begin{equation}
  \label{eq:opt_problem_epswise_l2_reg}
  \underset{E_1, \dots, E_N, A, b}{\text{minimize}} \ \lambda \left(\norm{E_1}_\text{fro}^2 +
    \dots + \norm{E_N}_\text{fro}^2 + \norm{A}_\text{fro}^2 \right)
  + \frac{1}{M} \sum_{m=1}^M -\ln p\left( y^{(m)} \mid X^{(m)}\right).
\end{equation}

We calculate the objective's gradient with respect to the model's parameters using
backpropagation via Pytorch \citep{pytorch} autograd. We train the model in iterations,
periodically evaluating it on the validation dataset, and take the model with the best
validation accuracy as the final output of the training process.

\section{Experiments}
\label{sec:experiments}
\subsection{MNIST}

We tested DCTN with one EPS, \(\nu=0.5\) in \cref{eq:phi_definition}, \(K_1=4, Q_1=4\),
\(\text{lr} = 3 \cdot 10^{-3}\), \(\lambda = 0\) in \cref{eq:opt_problem_whole_tn_l2_reg},
batch size 128 on MNIST dataset with 50000/10000/10000 training/validation/test split. We got
98.75\% test accuracy. MNIST is considered relatively easy and doesn't represent modern
computer vision tasks \citep{fashionmnist_readme}.

\subsection{FashionMNIST}

FashionMNIST \citep{xiao2017fashion} is a dataset fully compatible with MNIST: it contains
70000 grayscale \(28 \times 28\) images. Each image belongs to one of 10 classes of
clothes. We split 70000 images into 50000/10000/10000
training/validation/test split and experimented with models with one, two, and three EPSes. The
more EPSes we used, the more overfitting DCTN experienced and the worse validation accuracy
got, so we didn't experiment with more than three EPSes.
For one, two, and three EPSes, we chose hyperparameters by a combination of gridsearch and
manual choosing and presented the best result (chosen by validation accuracy before being
evaluated on the test dataset) in \Cref{table:fashionmnist_experiments}.
In \Cref{sec:how_hyperparameters_affect_optimization_and_generalization}, we describe
more experiments and discuss how various hyperparameters affect optimization and generalization of DCTN.

\begin{table}[h]
% see projects/dctn/small_experiments/pre_publishing_test_evaluation
  \caption{Comparison of our best models (top 3 rows) with 1, 2, and 3 EPSes, respectively,
    with the best (by a combination of accuracy and parameter count) existing models on
    FashionMNIST dataset. DCTN with one EPS wins against existing models with similar parameter
    count. Adding more EPSes makes test accuracy worse due to overfitting. All 3 of our models
    eventually reach nearly 100\% accuracy if not stopped early. We trained all DCTNs
    with batch size 128.} % Should I remove values of hyperparameters here?
\label{table:fashionmnist_experiments}
\resizebox{\textwidth}{!}{%
\begin{tabular}{p{13cm}lr}
  Model & Accuracy & Parameter count \\

  \hline One EPS \(K_1{=}4\), \(Q_1{=}4\), \(\nu{=}0.5\),
  \(E_1 {\sim} \mathcal{N}(\mu{=}0, \sigma{=}0.25)\),
  \(A,b {\sim} U[-(H_1 W_1 Q_1)^{-0.5}, -(H_1 W_1 Q_1)^{0.5}]\),
  \(\text{lr}{=}3\cdot10^{-3}\), \(\lambda{=}0\), \cref{eq:opt_problem_whole_tn_l2_reg}
        & 89.38\% & \(2.9 \cdot 10^5\) \\
        % eps_plus_linear_fashionmnist/replicate_90.19_vacc/2020-05-04T23:13:52_stopped_manually

  \hline Two EPSes, \(K_1{=}4, Q_1{=}4, K_2{=}3, Q_2{=}6\), \(\nu {\approx} 1.46\),
  EPSes initialized from \(\mathcal{N}(\mu{=}0, \sigma{=}Q_\text{in}^{-0.5 K^2})\),
  \(A {\sim} \mathcal{N}(\mu{=}0, \sigma{=}0.25(H_2 W_2 Q_2)^{-0.5})\),
  \(b {\sim} U[-(H_2 W_2 Q_2)^{-0.5}, (H_2 W_2 Q_2)^{-0.5}]\),
  \(\text{lr}{=}1.11\cdot10^{-4}\),
  \(\lambda{=}10^{-2}\), \cref{eq:opt_problem_epswise_l2_reg} & 87.65\% & \(1.8 \cdot 10^6\) \\ % 2_epses_plus_linear/adam_and_epswise_l2/2020-04-26T23:06:19_earlystopped

  \hline Three EPSes, \(K_1{=}4, Q_1{=}4, K_2{=}3, Q_2{=}12, K_2{=}2, Q_2{=}24\),
  \(\nu {\approx} 1.46\), EUSIR initialization of EPSes (see \Cref{sec:initialization_and_scaling_of_input}),
  \(A {\sim} \mathcal{N}(\mu{=}0, \sigma{=}0.25 (H_2 W_2 Q_2)^{-0.5})\),
  \(b {\sim} U[-(H_3 W_3 Q_3)^{-0.5}, (H_3 W_3 Q_3)^{-0.5}]\),
  \(\text{lr}{=}10^{-7}\), \(\lambda {=} 10^{-1}\), \cref{eq:opt_problem_whole_tn_l2_reg}
        & 75.94\% & \(4\cdot10^6\) \\ % 3_epses_plus_linear_fashionmnist/2020-05-12T19:33:11_vacc=0.7708_manually_stopped
%\hline DENSER ensembling                                                                & 95.26\%  & ?               \\
\hline GoogleNet + Linear SVC                                                           &
                                                                                          93.7\%   & \(6.8\cdot 10^6\)           \\
\hline VGG16                                                                            &
                                                                                          93.5\%
                   & \(2.6 \cdot 10^7\)           \\
\hline CNN: 5x5 conv -\textgreater 5x5 conv -\textgreater linear -\textgreater linear   &
                                                                                          91.6\%
                   & \(3.3 \cdot 10^6\)           \\
\hline AlexNet + Linear SVC                                                             &
                                                                                          89.9\%
                   & \(6.2 \cdot 10^7\)           \\
\hline Matrix tensor train in snake pattern (Glasser 2019)                              & 89.2\%   & ?               \\
\hline Multilayer perceptron                                                            &
                                                                                          88.33\%
                   & \(2.3 \cdot 10^5\)          
\end{tabular}}
\end{table}

\subsection{CIFAR10}

CIFAR10~\citep{cifar10} is a colored dataset of 32 by 32 images in 10 classes. We used 45000/5000/10000 train/validation/test split. We evaluated DCTN on the colored version using YCbCr color scheme and on grayscale version which mimics MNIST and FashionMNIST. The results are in \Cref{table:cifar10_experiments}.
DCTN overfits and performs poorly -- barely better than a linear classifier. Our hypotheses for why DCTN performs poorly on CIFAR10 in contrast to MNIST and FashionMNIST are:
(a)~CIFAR10 images have much less zero values; (b)~classifying CIFAR10 is a much more difficult problem; (c)~making CIFAR10 grayscale loses too much useful information, while non-grayscale version has too many features, which leads to overfitting. In the future work, we are going to check these hypotheses with intensive numerical experiments.

\begin{table}[h]
% see https://github.com/philip-bl/dctn
  \caption{DCTN results on CIFAR10. For each number of color channels, for each number of EPSes, we chose the
    kernel sizes \(K_n\), the quantum dimension sizes \(Q_n\), and the learning rate using grid
    search (excluding models the training of which didn't fit in 8 Gb of videocard's RAM) and
    showed the best model in the table. All of
    these models can reach almost 100\% training accuracy if not stopped early. Two bottom rows
    show the accuracy of a linear classifier and of one of the state of the art CNNs for comparison.}
\label{table:cifar10_experiments}
\centering
\begin{tabular}{ccc}
Channels & Model & Accuracy\\
\hline Grayscale & One EPS, \(K_1{=}4, Q_1{=}4\) & 49.5\%\\ % https://github.com/philip-bl/dctn#dctn-using-grayscale-cifar10
\hline Grayscale & Two EPSes, \(K_1{=}4, Q_1{=}4, K_2{=}3, Q_2{=}6\) & 54.8\%\\ % https://github.com/philip-bl/dctn#dctn-using-grayscale-cifar10
\hline YCbCr & One EPS, \(K_1{=}2, Q_1{=}24\) & 51\%\\ % https://github.com/philip-bl/dctn#dctn-using-ycbcr
\hline YCbCr & Two EPSes, \(K_1{=}2, Q_1{=}23, K_2{=}2, Q_2{=}24\) & 38.6\%\\ % dctn/small_experiments/plots/12_cifar10_ycbcr_plus_constant_channel_two_epses_K=2_gridsearch/all_experiments.html
\hline RGB & Linear classifier & 41.73\%\\ % https://github.com/philip-bl/dctn#baseline-linear-classifier
\hline RGB & EfficientNet-B7~\citep{efficientnet} & 98.9\%\\ % 
\end{tabular}
\end{table}
% here I describe the best experiments I've got on FashionMNIST

\section{Conclusion}

We showed that a tensor regression model can have locality, parameter sharing, and depth,
just like a neural network. To check if such models are promising, we showed how to implement
EPS as a backpropagatable function/layer and built a novel tensor regression model called DCTN
consisting of a composition of EPSes. In principle, DCTN can be used for tasks for which deep
CNNs are used.

We tested it for image classification on MNIST, FashionMNIST, and CIFAR10 datasets. We found
that shallow DCTN with one EPS performs well on MNIST and FashionMNIST while having a small
parameter.
Unfortunately, adding more EPSes increased overfitting and thus made
accuracy worse. Moreover, DCTN performed very badly on CIFAR10 regardless of depth. This
suggests we can't straightforwardly copy characteristics of NNs to make tensor regression work
better. We think that overfitting is a large problem for tensor regression. In
\Cref{sec:how_hyperparameters_affect_optimization_and_generalization}, we have discussed how
hyperparameters, such as initialization, input preprocessing, and learning rate affect
optimization and overfitting. It seems that these hyperparameters have a large effect, but our
understanding of this is limited. For example, we got our best model by scaling down the
multiplier \(\nu\) used in the input preprocessing function \cref{eq:phi_definition}. We think
it's important to study the effects of hyperparameters further and understand why some of them
help fight overfitting.

Our code is free software and can be accessed at
%\emph{anonymized url}.
\url{https://github.com/philip-bl/dctn}.

\section*{Acknowledgements}
The work of Anh-Huy Phan was supported by the Ministry of Education
and Science of the Russian Federation under Grant 14.756.31.0001.

\bibliographystyle{plainnat}
\bibliography{bibliography}

\begin{thebibliography}{28}
\providecommand{\natexlab}[1]{#1}
\providecommand{\url}[1]{\texttt{#1}}
\expandafter\ifx\csname urlstyle\endcsname\relax
  \providecommand{\doi}[1]{doi: #1}\else
  \providecommand{\doi}{doi: \begingroup \urlstyle{rm}\Url}\fi

\bibitem[Alammar(2018)]{illustrated_transformer}
Alammar.
\newblock The illustrated transformer.
\newblock \url{https://jalammar.github.io/illustrated-transformer}, 2018.
\newblock Accessed: 2020-09-30.

\bibitem[Bhatia et~al.(2019)Bhatia, Saggi, Kumar, and Jain]{bhatia2019matrix}
Amandeep~Singh Bhatia, Mandeep~Kaur Saggi, Ajay Kumar, and Sushma Jain.
\newblock Matrix product state--based quantum classifier.
\newblock \emph{Neural computation}, 31\penalty0 (7):\penalty0 1499--1517,
  2019.

\bibitem[Biamonte and
  Bergholm(2017)]{biamonte2017tensor_networks_in_a_nutshell}
Jacob Biamonte and Ville Bergholm.
\newblock Tensor networks in a nutshell.
\newblock \emph{arXiv preprint arXiv:1708.00006}, 2017.

\bibitem[Bridgeman and
  Chubb(2017)]{bridgeman2017handwaving_and_interpretive_dance}
Jacob~C Bridgeman and Christopher~T Chubb.
\newblock Hand-waving and interpretive dance: an introductory course on tensor
  networks.
\newblock \emph{Journal of Physics A: Mathematical and Theoretical},
  50\penalty0 (22):\penalty0 223001, 2017.

\bibitem[Cichocki et~al.(2017)Cichocki, Phan, Oseledets, Zhao, Sugiyama, Lee,
  and Mandic]{cichocki_part_2}
A~Cichocki, AH~Phan, I~Oseledets, Q~Zhao, M~Sugiyama, N~Lee, and D~Mandic.
\newblock Tensor networks for dimensionality reduction and large-scale
  optimizations: Part 2 applications and future perspectives.
\newblock \emph{Foundations and Trends in Machine Learning}, 9\penalty0
  (6):\penalty0 431--673, 2017.

\bibitem[Cichocki et~al.(2016)Cichocki, Lee, Oseledets, Phan, Zhao, and
  Mandic]{cichocki2016part1}
Andrzej Cichocki, Namgil Lee, Ivan~V Oseledets, A-H Phan, Qibin Zhao, and
  D~Mandic.
\newblock Low-rank tensor networks for dimensionality reduction and large-scale
  optimization problems: Perspectives and challenges part 1.
\newblock \emph{arXiv preprint arXiv:1609.00893}, 2016.

\bibitem[Cohen et~al.(2016)Cohen, Sharir, and Shashua]{cohen2016expressive}
Nadav Cohen, Or~Sharir, and Amnon Shashua.
\newblock On the expressive power of deep learning: A tensor analysis.
\newblock In \emph{Conference on Learning Theory}, pages 698--728, 2016.

\bibitem[Ehrbar(2000)]{ehrbar2000graph_notation_for_arrays}
Hans~G Ehrbar.
\newblock Graph notation for arrays.
\newblock \emph{ACM SIGAPL APL Quote Quad}, 31\penalty0 (3):\penalty0 13--26,
  2000.

\bibitem[{Glasser} et~al.(2020){Glasser}, {Pancotti}, and
  {Cirac}]{glasser2018probabilistic}
I.~{Glasser}, N.~{Pancotti}, and J.~I. {Cirac}.
\newblock From probabilistic graphical models to generalized tensor networks
  for supervised learning.
\newblock \emph{IEEE Access}, 8:\penalty0 68169--68182, 2020.

\bibitem[Grant et~al.(2018)Grant, Benedetti, Cao, Hallam, Lockhart, Stojevic,
  Green, and Severini]{grant2018hierarchical}
Edward Grant, Marcello Benedetti, Shuxiang Cao, Andrew Hallam, Joshua Lockhart,
  Vid Stojevic, Andrew~G Green, and Simone Severini.
\newblock Hierarchical quantum classifiers.
\newblock \emph{npj Quantum Information}, 4\penalty0 (1):\penalty0 1--8, 2018.

\bibitem[He et~al.(2015)He, Zhang, Ren, and Sun]{he_initialization}
Kaiming He, Xiangyu Zhang, Shaoqing Ren, and Jian Sun.
\newblock Delving deep into rectifiers: Surpassing human-level performance on
  imagenet classification.
\newblock In \emph{Proceedings of the IEEE international conference on computer
  vision}, pages 1026--1034, 2015.

\bibitem[Huggins et~al.(2019)Huggins, Patil, Mitchell, Whaley, and
  Stoudenmire]{huggins2019towards}
William Huggins, Piyush Patil, Bradley Mitchell, K~Birgitta Whaley, and E~Miles
  Stoudenmire.
\newblock Towards quantum machine learning with tensor networks.
\newblock \emph{Quantum Science and technology}, 4\penalty0 (2):\penalty0
  024001, 2019.

\bibitem[Karpathy(2019)]{karpathy_recipe}
Karpathy.
\newblock A recipe for training neural networks.
\newblock \url{https://karpathy.github.io/2019/04/25/recipe}, 2019.
\newblock Accessed: 2020-05-25.

\bibitem[Kingma and Ba(2014)]{kingma2014adam}
Diederik~P Kingma and Jimmy Ba.
\newblock Adam: A method for stochastic optimization.
\newblock \emph{arXiv preprint arXiv:1412.6980}, 2014.

\bibitem[Krizhevsky et~al.(2009)Krizhevsky, Hinton, et~al.]{cifar10}
Alex Krizhevsky, Geoffrey Hinton, et~al.
\newblock Learning multiple layers of features from tiny images.
\newblock 2009.

\bibitem[Liu et~al.(2019)Liu, Ran, Wittek, Peng, Garc{\'\i}a, Su, and
  Lewenstein]{liu2019ml_by_unitary_tn_of_hierarchical_tree_structure}
Ding Liu, Shi-Ju Ran, Peter Wittek, Cheng Peng, Raul~Bl{\'a}zquez Garc{\'\i}a,
  Gang Su, and Maciej Lewenstein.
\newblock Machine learning by unitary tensor network of hierarchical tree
  structure.
\newblock \emph{New Journal of Physics}, 21\penalty0 (7):\penalty0 073059,
  2019.

\bibitem[Miller(2019)]{torchmps}
Jacob Miller.
\newblock Torchmps.
\newblock \url{https://github.com/jemisjoky/torchmps}, 2019.

\bibitem[Miller et~al.(2020)Miller, Rabusseau, and
  Terilla]{miller_2020_umps_psm}
Jacob Miller, Guillaume Rabusseau, and John Terilla.
\newblock Tensor networks for probabilistic sequence modeling, 2020.

\bibitem[Novikov et~al.(2016)Novikov, Trofimov, and
  Oseledets]{novikov2016exponential}
Alexander Novikov, Mikhail Trofimov, and Ivan Oseledets.
\newblock Exponential machines.
\newblock \emph{arXiv preprint arXiv:1605.03795}, 2016.

\bibitem[Or{\'u}s(2014)]{orus2014practical_introduction_to_tns}
Rom{\'a}n Or{\'u}s.
\newblock A practical introduction to tensor networks: Matrix product states
  and projected entangled pair states.
\newblock \emph{Annals of Physics}, 349:\penalty0 117--158, 2014.

\bibitem[PapersWithCode(2020)]{paperswithcode_sota}
PapersWithCode.
\newblock Browse the state-of-the-art in machine learning.
\newblock \url{https://paperswithcode.com/sota}, 2020.
\newblock Accessed: 2020-05-23.

\bibitem[Paszke et~al.(2019)Paszke, Gross, Massa, Lerer, Bradbury, Chanan,
  Killeen, Lin, Gimelshein, Antiga, Desmaison, Kopf, Yang, DeVito, Raison,
  Tejani, Chilamkurthy, Steiner, Fang, Bai, and Chintala]{pytorch}
Adam Paszke, Sam Gross, Francisco Massa, Adam Lerer, James Bradbury, Gregory
  Chanan, Trevor Killeen, Zeming Lin, Natalia Gimelshein, Luca Antiga, Alban
  Desmaison, Andreas Kopf, Edward Yang, Zachary DeVito, Martin Raison, Alykhan
  Tejani, Sasank Chilamkurthy, Benoit Steiner, Lu~Fang, Junjie Bai, and Soumith
  Chintala.
\newblock Pytorch: An imperative style, high-performance deep learning library.
\newblock In H.~Wallach, H.~Larochelle, A.~Beygelzimer, F.~d\textquotesingle
  Alch\'{e}-Buc, E.~Fox, and R.~Garnett, editors, \emph{Advances in Neural
  Information Processing Systems 32}, pages 8024--8035. Curran Associates,
  Inc., 2019.
\newblock URL
  \url{http://papers.neurips.cc/paper/9015-pytorch-an-imperative-style-high-performance-deep-learning-library.pdf}.

\bibitem[Polu and Sutskever(2020)]{gpt_f}
Stanislas Polu and Ilya Sutskever.
\newblock Generative language modeling for automated theorem proving, 2020.

\bibitem[Robeva and Seigal(2019)]{robeva2019duality}
Elina Robeva and Anna Seigal.
\newblock Duality of graphical models and tensor networks.
\newblock \emph{Information and Inference: A Journal of the IMA}, 8\penalty0
  (2):\penalty0 273--288, 2019.

\bibitem[Stoudenmire and Schwab(2016)]{stoudenmire1605supervised}
E.~Miles Stoudenmire and David~J. Schwab.
\newblock Supervised learning with quantum-inspired tensor networks, 2016.

\bibitem[Tan and Le(2020)]{efficientnet}
Mingxing Tan and Quoc~V. Le.
\newblock Efficientnet: Rethinking model scaling for convolutional neural
  networks, 2020.

\bibitem[Xiao et~al.(2017{\natexlab{a}})Xiao, Rasul, and
  Vollgraf]{fashionmnist_readme}
Han Xiao, Kashif Rasul, and Roland Vollgraf.
\newblock Fashionmnist readme.
\newblock
  \url{https://github.com/zalandoresearch/fashion-mnist/blob/master/README.md},
  2017{\natexlab{a}}.
\newblock Accessed: 2020-05-24.

\bibitem[Xiao et~al.(2017{\natexlab{b}})Xiao, Rasul, and
  Vollgraf]{xiao2017fashion}
Han Xiao, Kashif Rasul, and Roland Vollgraf.
\newblock Fashion-mnist: a novel image dataset for benchmarking machine
  learning algorithms.
\newblock \emph{arXiv preprint arXiv:1708.07747}, 2017{\natexlab{b}}.

\end{thebibliography}

\clearpage
\appendix

\section{How hyperparameters affect optimization and generalization}
\label{sec:how_hyperparameters_affect_optimization_and_generalization}

In this section, we present some of our thoughts and findings about how the
hyperparameters affect DCTN's optimization and overfitting. Hopefully, they may prove useful
for other tensor regression models as well. All empirical findings presented here were obtained
on FashionMNIST dataset. They might not replicate on other datasets or with radically different
values of hyperparameters.

\subsection{Initialization of the model and scaling of the input}
\label{sec:initialization_and_scaling_of_input}
Consider a DCTN with \(N\) EPSes and kernel sizes \(K_1, \dots, K_N\). Let \(c > 0\) be a
positive real number. Then
% \ifdefined\PHILIPSTHESIS
\begin{equation}
  A \cdot \operatorname{vec}\left(
    \operatorname{eps}(E_N, \dots \operatorname{eps}(E_1, c X_0) \dots )
  \right) =
  c^{K_1^2 \dots K_N^2} A \cdot \operatorname{vec}\left(
    \operatorname{eps}(E_N, \dots \operatorname{eps}(E_1, X_0) \dots )
  \right)
  \label{eq:multiplyingx0byc}
\end{equation}
and
\begin{equation}
  A \cdot \operatorname{vec}\left(
    \operatorname{eps}(E_N, \dots \operatorname{eps}(c E_1, X_0) \dots )
  \right) =
  c^{K_2^2 \dots K_N^2} A \cdot \operatorname{vec}\left(
    \operatorname{eps}(E_N, \dots \operatorname{eps}(E_1, X_0) \dots )
  \right).
  \label{eq:multiplyingepsbyc}
\end{equation}
% \else
% \begin{equation}
%   A \cdot \operatorname{vec}\left(
%     \operatorname{eps}(E_N, \dots \operatorname{eps}(E_2, \operatorname{eps}(E_1, c X_0)) \dots )
%   \right) =
%   c^{K_1^2 K_2^2 \dots K_N^2} A \cdot \operatorname{vec}\left(
%     \operatorname{eps}(E_N, \dots \operatorname{eps}(E_2, \operatorname{eps}(E_1, X_0)) \dots )
%   \right)
%   \label{eq:multiplyingx0byc}
% \end{equation}
% and
% \begin{equation}
%   A \cdot \operatorname{vec}\left(
%     \operatorname{eps}(E_N, \dots \operatorname{eps}(E_2, \operatorname{eps}(c E_1, X_0)) \dots )
%   \right) =
%   c^{K_2^2 \dots K_N^2} A \cdot \operatorname{vec}\left(
%     \operatorname{eps}(E_N, \dots \operatorname{eps}(E_2, \operatorname{eps}(E_1, X_0)) \dots )
%   \right).
%   \label{eq:multiplyingepsbyc}
% \end{equation}  
% \fi

Since \(K_1^2 K_2^2 \dots K_N^2\) can get very large (e.g. 576 for \(K_1=4, K_2=3, K_3=2\)), it
follows that if the constant \(\nu\) in the input preprocessing function
\begin{equation}
  \varphi(x) = \nu \begin{bmatrix}\cos^2(\frac{\pi}{2} x) \\ \sin^2(\frac{\pi}{2}
    x)\end{bmatrix}  \tag{\ref{eq:phi_definition}}
\end{equation}
is chosen slightly larger than optimal, \cref{eq:multiplyingx0byc} might easily get infinities
in floating point arithmetic, and if it's chosen slightly smaller than optimal,
\cref{eq:multiplyingx0byc} might easily become all zeros, and the model's output will stop
depending on anything except the bias \(b\).

The same is true in a slightly lesser degree for scaling of initial values of \(E_1, \dots,
E_N\), especially for the earlier EPSes, as shown in \cref{eq:multiplyingepsbyc}. Also, if for
a chosen \(\nu\) and chosen initialization of the EPSes, the values in
\[
  A \cdot \operatorname{vec}\left(
    \operatorname{eps}(E_N, \dots \operatorname{eps}(E_2, \operatorname{eps}(E_1, X_0)) \dots )
  \right)
\]
have large standard deviation, then the values in the output of the whole model
\[
  A \cdot \operatorname{vec}\left(
    \operatorname{eps}(E_N, \dots \operatorname{eps}(E_2, \operatorname{eps}(E_1, X_0)) \dots )
  \right) + b
\]
will have large standard deviation as well, which might lead to initial negative log likelihood being
high. \citep{karpathy_recipe} recommends initializing neural networks for classification in
such a way that initially the loss has the best
possible value given that your model is allowed to know the proportion of labels in the
datasets, but hasn't been allowed to train yet. For example, if you have 10 possible labels
with equal number of samples, a perfectly calibrated model that is ignorant about the images
should have negative log likelihood equal to \( \ln 10 \approx 2.3 \). We think that if the model
starts with negative log likelihood much higher than this value, problems with the optimization process might
occur.

One way we tried to overcome this difficulty was by adapting \emph{He initialization}
\citep{he_initialization} for EPSes:
\begin{equation}
  E \sim \mathcal{N}(\mu=0, \sigma=Q_\text{in}^{-0.5 K^2}).
  \label{eq:theoreticalinitofE}
\end{equation}
The rationale for this initialization is that if the components of
\(E \in \mathbb{R}^{Q_\text{out} \times Q_\text{in} \times \dots \times Q_\text{in}}\)
are distributed i.i.d. with zero mean and variance \(\alpha^2\), and if the components of
\(\omega \in \mathbb{R}^{Q_\text{in} \times \dots \times Q_\text{in}}\)
are distributed i.i.d. with mean \(\mu\) and variance \(\sigma^2\), then, applying the EPS
\(E\) similar to
\cref{eq:eps_output_one_pixel_matmul}, we have
\[ \mathbb{E}\left[\operatorname{mat}(E) \cdot \operatorname{vec}(\omega)\right] = 0,\]
\[ \operatorname{Var}\left[ \operatorname{mat}(E) \cdot \operatorname{vec}(\omega) \right] =
  Q_\text{in}^{K^2} \alpha^2 (\sigma^2 + \mu^2) I.\]

Note that the input \(\omega\) having i.i.d. coordinates is not necessarily true in the real scenario, but
still, we might try to initialize the EPSes using He initialization
\cref{eq:theoreticalinitofE}. In this case, we choose such value for \(\nu\) as to have the
components of the vector
\[
  \operatorname{vec} \left(
    \bigotimes_{\delta h = 0}^{K_1-1} \bigotimes_{\delta w = 0}^{K_1-1}
    X_0(h + \delta h, w + \delta w)
  \right),
\]
which appears in \cref{eq:eps_output_one_pixel_matmul}, have empirical mean \(\mu\) and
empirical standard deviation \(\sigma\) (over the whole training dataset) satisfy \(\mu^2 +
\sigma^2 = 1\). For example, on FashionMNIST with our choice of \(\varphi\), the value \(\nu
\approx 1.46\) satisfies this criterion, and that's the value we use in 2 out of 3 experiments
in \Cref{table:fashionmnist_experiments}.

However, empirically we've seen that with He initialization of a DCTN with 2 EPSes, the empirical
standard deviation (over the whole training dataset) of the second intermediate representation
\(\operatorname{std}(X_2)\) sometimes (depending on the random seed) is magnitudes larger or
smaller than \(1\). If it's large, this leads to initial negative log likelihood loss being
high, which we think might be bad for optimization. That's why we devised another
initialization scheme: while choosing \( \nu \) the same way described earlier, we first
initialize components of each EPS \(E_n\) from the standard normal distribution and then
rescale the EPS by the number required to make empirical standard deviation (over the whole
training dataset) of its output \(X_n\) equal to 1. In other words, here's what we do:
\begin{align*}
  & \text{Initialize } E_1 \sim \mathcal{N}(0, 1)\\
  & \text{Multiply } E_1 \text{ by the number that will make }
    \operatorname{std}(\operatorname{eps}(E_1, X_0)) = 1 \\
  & \text{Initialize } E_2 \sim \mathcal{N}(0, 1)\\
  & \text{Multiply } E_2 \text{ by the number that will make }
    \operatorname{std}(\operatorname{eps}(E_2, \operatorname{eps}(E_1, X_0))) = 1 \\
  & \vdots\\
  & \text{Initialize } E_N \sim \mathcal{N}(0, 1)\\
  & \text{Multiply } E_N \text{ by the number that will make } %
    \operatorname{std}(\operatorname{eps}(E_N, \dots \operatorname{eps}(E_1, X_0) \dots )) = 1
\end{align*}

\begin{figure}[h]
  \centering
  \begin{minipage}{0.45\columnwidth}
    \subfloat[]{\includegraphics[width=\columnwidth]{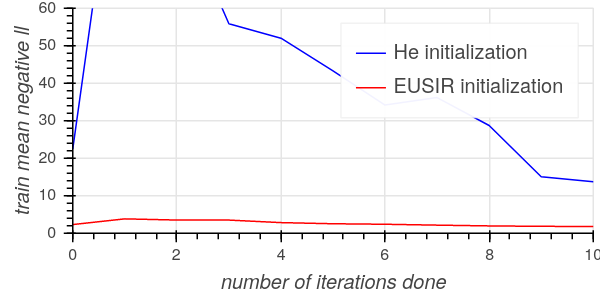}}
    
    \subfloat[]{\includegraphics[width=\columnwidth]{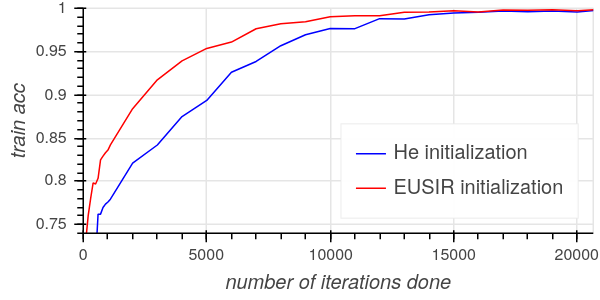}}
  \end{minipage}
  \begin{minipage}{0.5\columnwidth}
    \subfloat[]{\includegraphics[width=\columnwidth]{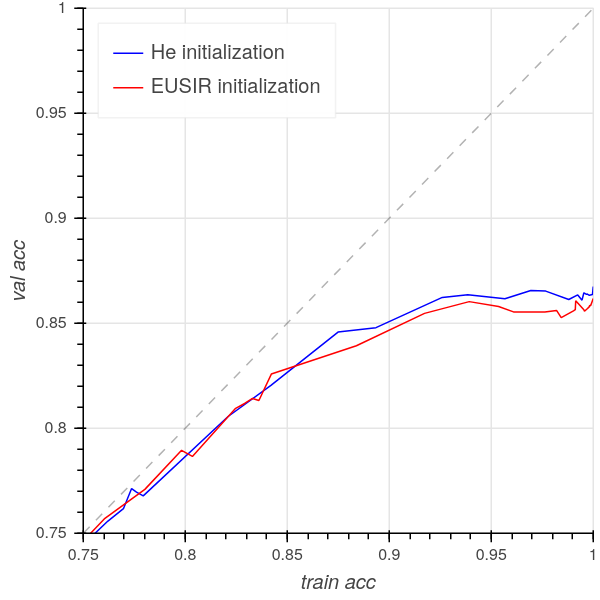}}
  \end{minipage}
  
  \caption[Comparison of He initialization and empirical unit std of intermediate
  representations initialization]{Comparison of He initialization and EUSIR initialization.
    (a) He initialization suffers from high initial loss.
    (b) EUSIR initialization trains faster.
    (c) Unfortunately, this didn't lead to less overfitting.
    In all three plots,
    the model consists of 2 EPSes with \(K_1{=}3,Q_1{=}4,
    K_2{=}3, Q_2{=}6,\text{lr}=4\cdot10^{-5},\lambda=10^{-2}\), and the objective function
    \cref{eq:opt_problem_whole_tn_l2_reg}.}
  \label{fig:he_vs_empirical}
\end{figure}

We call this
initialization scheme EUSIR initialization (\emph{empirical unit std of intermediate representations
  initialization}). You can see a visualization of its effects in \Cref{fig:he_vs_empirical}.

\subsection{Other hyperparameters}
\label{sec:other_hyperparameters}
\begin{figure}[h]
  \centering
  \begin{minipage}{0.48\columnwidth}
    \subfloat[]{\includegraphics[width=\columnwidth]{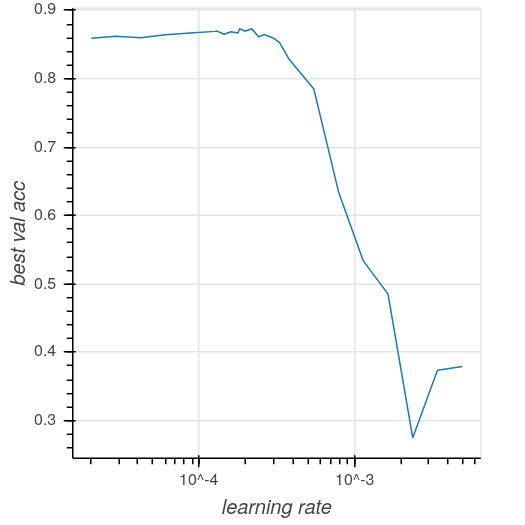}}
  \end{minipage}
  \begin{minipage}{0.48\columnwidth}
    \subfloat[]{\includegraphics[width=\columnwidth]{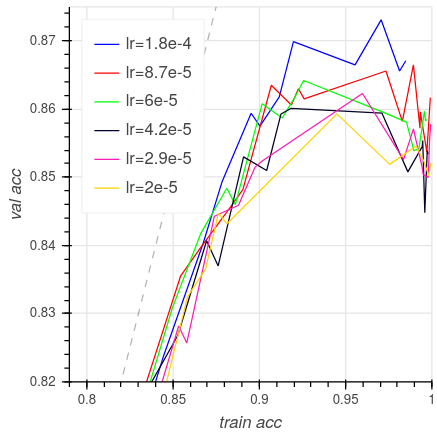}}
  \end{minipage}
  \caption[Effects of learning rate]{As can be seen in (a), too large learning rate causes
    training to not converge.  However (b) shows that large but not too large learning rate
    slightly reduces overfitting (training trajectories with higher learning rate achieve
    higher validation accuracy). This is in line with folk understanding of how learning rate
    affects training in deep learning. The model had 2 EPSes with
    \(K_1{=}4, Q_1{=}4, K_2{=}3, Q_2{=}=6, \lambda=10^{-2}\), used the objective function
    \cref{eq:opt_problem_whole_tn_l2_reg} and EUSIR initialization.}
  \label{fig:how_lr_affects_everything}
\end{figure}

\begin{figure}[h]
  \centering
%  \begin{minipage}{0.48\columnwidth}
    \subfloat[]{\includegraphics[width=0.48\textwidth]{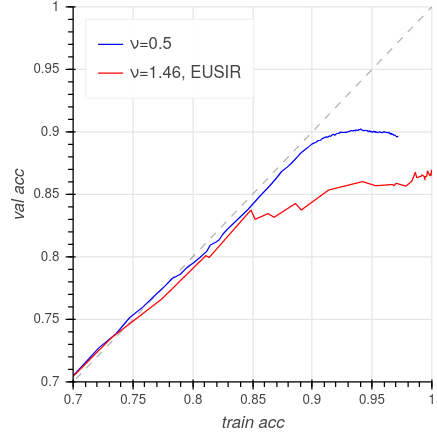}}
%  \end{minipage}
%  \begin{minipage}{0.48\columnwidth}
    \subfloat[]{\includegraphics[width=0.48\textwidth]{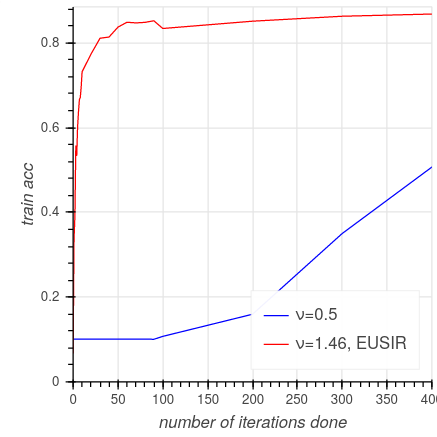}}
%  \end{minipage}
  \caption[Accidentally chosen hyperparameters led to better generalization
  but slower training]{The plot (a) shows how a model we made by accidentally altering hyperparameters
    achieves better generalization. However, it trained 960 times slower, as can be seen in
    (b). Both models have one
    EPS with \(K_1{=}4, Q_1{=}4, \text{lr}{=}3 \cdot 10^{-3}, \lambda=0\),
    \(A,b \sim U[-0.02, 0.02]\). The difference is in
    the choice of \(\nu\) and in initialization of EPSes. The blue model is the model from the
    first
    row of \Cref{table:fashionmnist_experiments}. It has \(\nu{=}0.5\) and its intermediate
    representations have standard deviations
    \(\operatorname{std}(X_1){\approx}1.7 \cdot 10^{-6}\),
    \(\operatorname{std}(A \cdot \operatorname{vec}(X_1)){\approx}1.1 \cdot 10^{-6}\).
    The red model uses \(\nu{\approx}1.46\), EUSIR initialization and thus has,
    \(\operatorname{std}(X_1){=}1\). Notice that the standard deviation of the output of linear
    layer of the blue model, if we don't add the bias \(b\), is very small compared to the
    standard deviation of the bias \(b\). We speculate that
    this is probably the reason of much better generalization. It
    important to understand why the blue model's initialization and the choice of
    \(\nu\) worked so well and figure out how to achieve it with more EPSes.}
  \label{fig:one_eps_small_nu_vs_empirical}
\end{figure}

\begin{itemize}
\item \Cref{fig:how_lr_affects_everything} discusses how high learning rate leads to less
  overfitting.
\item \Cref{fig:one_eps_small_nu_vs_empirical} discusses how we accidentally got the best
  result with one EPS by setting a very small \(\nu\).
\item In our experiments, \(\ell_2\) regularization coefficient \(\lambda\) affected neither
  training speed nor validation accuracy. We don't provide plots depicting this, because they
  would show nearly identical training trajectories for different values of \(\lambda\) from
  \(0\) to \(10^{-2}\).
\end{itemize}

% \appendix
% \section{Representing EPS in tensor formats}
% Here I describe an empirically untested idea of representing EPS in Snake MTT
% format or in CP format. This will regularize it and allow us to have larger EPS.
% \input{representing_eps_in_tensor_formats}

% \section{Initialization similar to residual connection}
% Here I describe it. Maybe I don't actually need this section, who knows. Well, it's
% definitely not number 1 priority.

\end{document}